\documentclass[
]{ceurart}

\usepackage{listings}
\usepackage{hyperref}       
\usepackage{url}            
\usepackage{booktabs}       
\usepackage{amsfonts}       
\usepackage{nicefrac}       
\usepackage{microtype}      
\usepackage{xcolor}         

\usepackage{amsmath}
\usepackage[LSBC5,LSBC3,LSBC4,LSBC2,LSBC1,T1]{fontenc}  
\usepackage{chessboard}

 \usepackage{xskak}
 \usepackage{subcaption}
 \usepackage{cleveref}

\newcommand{\revision}[1]{{#1}}

\begin{document}

\conference{}

\title{Towards Piece-by-Piece Explanations for Chess Positions with SHAP}

\tnotetext[1]{To appear in CEUR-WS Proceedings.}

\author[1]{Francesco Spinnato}[%
orcid=0000-0002-3203-6716,
email=francesco.spinnato@unipi.it,
url=https://fspinna.github.io/,
]
\address[1]{University of Pisa, Largo B. Pontecorvo, 3, 56127, Pisa, Italy}

\begin{abstract}
  Contemporary chess engines offer precise yet opaque evaluations, typically expressed as centipawn scores. While effective for decision-making, these outputs obscure the underlying contributions of individual pieces or patterns. In this paper, we explore adapting SHAP (SHapley Additive exPlanations) to the domain of chess analysis, aiming to attribute a chess engine’s evaluation to specific pieces on the board. By treating pieces as features and systematically ablating them, we compute additive, per-piece contributions that explain the engine’s output in a locally faithful and human-interpretable manner. This method draws inspiration from classical chess pedagogy, where players assess positions by mentally removing pieces, and grounds it in modern explainable AI techniques. Our approach opens new possibilities for visualization, human training, and engine comparison. We release accompanying code and data to foster future research in interpretable chess AI.
\end{abstract}

\begin{keywords}
  chess \sep
  explainable AI \sep
  shap
\end{keywords}

\maketitle

\section{Introduction}
Evaluating a chess position is a complex endeavor that combines long-term strategic foresight with immediate tactical precision. Contemporary chess engines summarize their assessments into a single scalar metric, typically expressed in centipawns, approximating the material advantage. This evaluation is indispensable for decision-making and training, yet it remains opaque: it does not reveal which specific positional elements underlie the overall judgment~\cite{mucke2022check}. This lack of interpretability poses challenges for human players who seek strategic clarity, as well as for researchers striving to understand the internal logic of modern engines~\cite{hammersborg2024information}.

In contrast, the field of explainable AI (XAI) in machine learning has developed a rich array of methods for interpreting model outputs in classification and regression tasks~\cite{bodria2023benchmarking}. Approaches such as feature attribution~\cite{ribeiro2016should}, saliency maps, and Shapley value decomposition~\cite{lundberg2017unified} have proven effective in explaining model decisions across various data modalities, including tabular~\cite{bodria2023benchmarking}, image~\cite{van2022explainable}, and time series data~\cite{poggioli2023text,spinnato2024fast,pludowski2025mascots}. Notably, many of these techniques are designed to be both model-agnostic and locally faithful, allowing them to be applied to any black-box model and to explain its behavior on a per-instance basis, providing actionable insights into complex decision-making and in critical domains~\cite{hulsen2023explainable,spinnato2022explaining,bianchi2024multivariate}. Recent work has started to adapt these methods to games, particularly chess, by focusing on high-level features such as material balance and king safety~\cite{palsson2023unveilingconcepts,czech2024representation}. However, current approaches have yet to deliver fine-grained, per-piece explanations that are simultaneously additive, position-specific, and grounded in rigorous attribution theory.

In this paper, we introduce an interpretability framework based on SHAP (SHapley Additive exPlanations)~\cite{lundberg2017unified}, a principled XAI method grounded in cooperative game theory. We treat individual pieces as features and quantify their contributions by systematically ablating them, removing each piece in turn, and measuring the resulting shift in evaluation. These perturbations define a local neighborhood of similar board states, from which SHAP derives additive attributions that reflect not only the standalone value of each piece but also its interaction with others.

Although removing pieces is not a legal operation within the rules of chess, the conceptual act of piece ablation has long been a tool in strategic analysis. Capablanca emphasized the value of simplification in order to clarify and exploit structural features, such as a pawn majority, that may become decisive in the endgame~\cite{capablanca1921chess}. Authors such as Dvoretsky~\cite{dvoretsky2023dvoretsky} and De la Villa~\cite{delavilla2024mistakes} have explored this idea as both a pedagogical and analytical approach, encouraging players to mentally remove pieces to clarify the strategic essence of a position. This practice naturally invites evaluative questions such as: ``What would happen if this piece were not on the board? Would my position improve?''. Beyond evaluation, such simplification can also enhance a player’s tactical recognition by revealing the most important pieces in the position. We aim to answer these questions with algorithmic precision, providing a systematic way to quantify the positional significance of each piece through principled feature attribution.
By grounding interpretability in the well-established framework of SHAP, our method enables new forms of analysis and visualization. It can clarify the roles of individual pieces, help commentators explain sudden shifts in engine evaluations, and guide training tools in emphasizing the most critical components of a position. Furthermore, this methodology can be used to compare engines, revealing differences in how neural or hybrid systems assign value across structurally similar positions.

In summary, we propose a SHAP-based interpretability method for chess engine evaluations that leverages robust XAI methodology from machine learning and adapts it to the structured, combinatorial nature of chess. Specifically, our contributions are as follows: \textit{\textbf{(1)}} We adapt SHAP to structured chess positions via a perturbation protocol based on piece ablation. \textit{\textbf{(2)}} We compute fine-grained, per-piece local attributions that decompose the evaluation into interpretable, additive contributions. \textit{\textbf{(3)}} We demonstrate with qualitative examples how these explanations can provide insights for several chess positional themes. \textit{\textbf{(4)}} We release an open-source implementation to foster reproducibility and encourage further research in interpretable chess AI\footnote{Code is available at: \url{https://github.com/fspinna/chessplainer}}. By aligning chess evaluation with mature techniques from explainable machine learning, our work opens a new direction for analyzing both the game and the algorithms that play it.

\section{Related Work}
Several recent studies have explored interpretability in chess engines from various perspectives. ~\cite{palsson2023unveilingconcepts} employed Shapley value sampling to attribute Stockfish’s evaluation output to a set of manually defined conceptual features, including “material,” “passed pawns,” and “king safety.” Their analysis analyzed how different evaluation pipelines, classical versus efficiently updateable neural networks (NNUE), weigh these high-level concepts. While this demonstrates the applicability of Shapley values to chess evaluation, the granularity remains semantic and abstract; individual pieces and their positions are not explicitly taken into account. Our work extends this line of inquiry to the level of concrete board elements, applying Shapley values directly to individual pieces to yield localized attributions for specific positions.
A complementary direction is explored in~\cite{gupta2023valuechess}, where the authors trained a neural network to estimate the marginal win probability associated with each piece-square pair. Their model captures general trends, such as the high utility of knights on central outposts, by learning a global value function over a large dataset. However, these estimates reflect statistical averages across many games and positions. In contrast, our method produces per-position explanations, allowing us to assess the situational importance of each piece with additive precision based on its context.

Other work has applied attribution tools from explainable AI to neural chess models. In~\cite{czech2024representation}, the authors analyze an AlphaZero-style transformer using Integrated Gradients (IG) to assess which input features drive predictions of win probability. The attribution focuses on feature channels, such as piece maps or auxiliary indicators, aggregated across datasets. While this highlights the utility of gradient-based techniques for model interpretation, it remains coarse in resolution. Our method offers finer granularity by assigning attribution directly to the pieces present on the board in a given position, rather than to broader input modalities.
A more perturbation-based approach is found in~\cite{puri2019explain}, which introduces SARFA (Specific and Relevant Feature Attribution) to explain move selection in chess and Go. SARFA identifies critical board squares by evaluating the impact of perturbations on move quality, balancing specificity with contextual relevance. Although SARFA and our method share a perturbation-based philosophy, their aims diverge: SARFA is designed to explain decisions (why a move was chosen), whereas we focus on evaluation decomposition (how each piece affects the position's value). 
Lastly,~\cite{kaushikan2024effects} examined the sensitivity of different engines to material and space advantage. Their results indicate that Stockfish places more weight on material factors than spatial ones, however, their analysis lacks the formal framework and guarantees provided by Shapley values. 

Taken together, these works highlight a growing interest in interpretability, including statistical analysis, neural network modeling, and attribution techniques, reflecting a broader trend toward understanding how complex evaluation functions derive their outputs. However, despite this momentum, none of the aforementioned methods apply Shapley-based techniques to estimate the local, per-piece contribution within specific board states via targeted ablation. Our method fills this gap by adapting SHAP to the structured domain of chess, producing additive, locally faithful explanations that enable engine transparency, instructional tools, and strategic insight.

\begin{figure}[t]
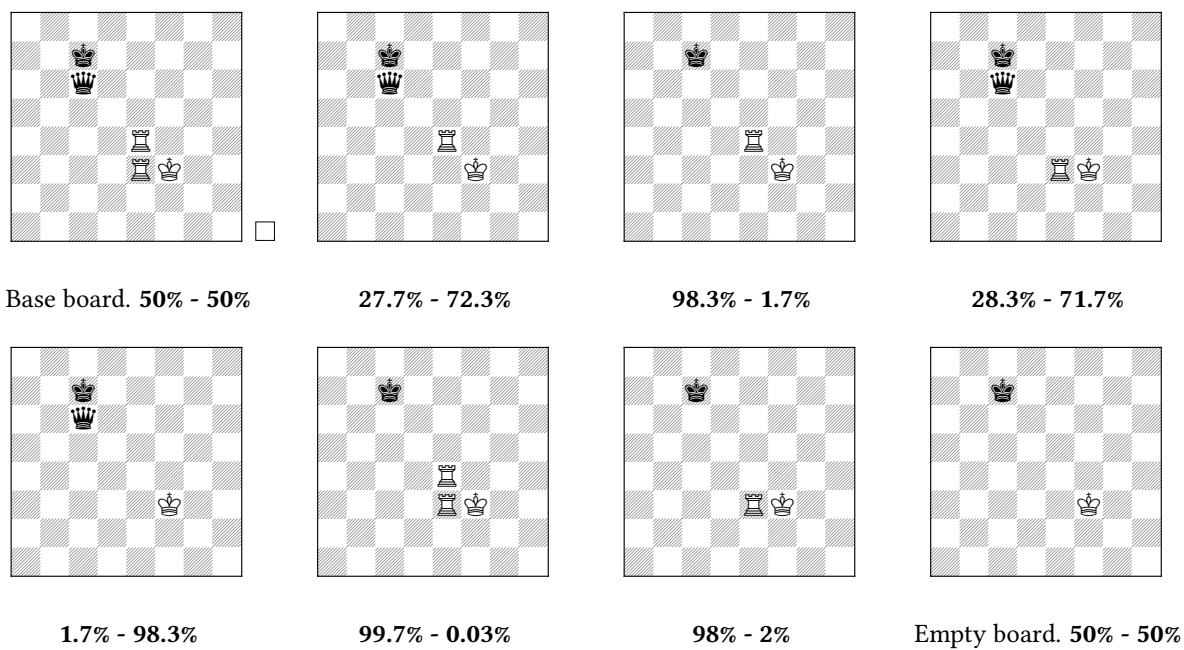

  \centering
  \begin{minipage}{\textwidth}
    \centering
    \begin{subfigure}[c]{0.24\linewidth}
      \centering
      \resizebox{\linewidth}{!}{%
        \newgame
        \chessboard[setfen=8/2k5/2q5/8/4R3/4RK2/8/8 w - - 0 1, showmover=true, label=false]
      }
      \caption*{Base board. \textbf{50\% - 50\%}}
    \end{subfigure}
    \hfill
    \begin{subfigure}[c]{0.24\linewidth}
      \centering
      \resizebox{\linewidth}{!}{%
        \newgame
        \chessboard[setfen=8/2k5/2q5/8/4R3/5K2/8/8 w - - 0 1, showmover=false, label=false]
      }
      \caption*{\textbf{27.7\% - 72.3\%}}
    \end{subfigure}
    \hfill
    \begin{subfigure}[c]{0.24\linewidth}
      \centering
      \resizebox{\linewidth}{!}{%
        \newgame
        \chessboard[setfen=8/2k5/8/8/4R3/5K2/8/8 w - - 0 1, showmover=false, label=false]
      }
      \caption*{\textbf{98.3\% - 1.7\%}}
    \end{subfigure}
    \hfill
    \begin{subfigure}[c]{0.24\linewidth}
      \centering
      \resizebox{\linewidth}{!}{%
        \newgame
        \chessboard[setfen=8/2k5/2q5/8/8/4RK2/8/8 w - - 0 1, showmover=false, label=false]
      }
      \caption*{\textbf{28.3\% - 71.7\%}}
    \end{subfigure}

    \begin{subfigure}[c]{0.24\linewidth}
      \centering
      \resizebox{\linewidth}{!}{%
        \newgame
        \chessboard[setfen=8/2k5/2q5/8/8/5K2/8/8 w - - 0 1, showmover=false, label=false]
      }
      \caption*{\textbf{1.7\% - 98.3\%}}
    \end{subfigure}
    \hfill
    \begin{subfigure}[c]{0.24\linewidth}
      \centering
      \resizebox{\linewidth}{!}{%
        \newgame
        \chessboard[setfen=8/2k5/8/8/4R3/4RK2/8/8 w - - 0 1, showmover=false, label=false]
      }
      \caption*{\textbf{99.7\% - 0.03\%}}
    \end{subfigure}
    \hfill
    \begin{subfigure}[c]{0.24\linewidth}
      \centering
      \resizebox{\linewidth}{!}{%
        \newgame
        
        \chessboard[setfen=8/2k5/8/8/8/4RK2/8/8 w - - 0 1, showmover=false, label=false]
      }
      \caption*{\textbf{98\% - 2\%}}
    \end{subfigure}
    \hfill
    \begin{subfigure}[c]{0.24\linewidth}
      \centering
      \resizebox{\linewidth}{!}{%
        \newgame
        \chessboard[setfen=8/2k5/8/8/8/5K2/8/8 w - - 0 1, showmover=false, label=false]
      }
      \caption*{Empty board. \textbf{50\% - 50\%}}
    \end{subfigure}
  \end{minipage}
  \caption{Perturbations generated by SHAP for evaluating the position on the top-left (white to move), evaluated using Lichess Win-Loss probability (\% white to win - \% black to win).}
  \label{fig:perturbations}
\end{figure}

\section{Explaining Chess Engines}

    \definecolor{shape3}{HTML}{E48065}
\definecolor{shape4}{HTML}{E58368}
\definecolor{shapc6}{HTML}{053061}

\begin{figure}[htbp]
  \centering
  \begin{subfigure}[c]{0.44\textwidth}
    \centering
  \resizebox{\linewidth}{!}{%
    \newgame
    \chessboard[
      setfen=8/2k5/2q5/8/4R3/4RK2/8/8 w - - 0 1,
      boardfontencoding=LSBC1,
      pgfstyle=color,
      opacity=0.7,
      color=shape3,
  colorbackfield={e3},
  color=shape4,
  colorbackfield={e4},
  color=shapc6,
  colorbackfield={c6}
    ]
  }
  \end{subfigure}%
  \hfill
  \begin{subfigure}[c]{0.55\textwidth}
    \centering
    \includegraphics[width=\linewidth]{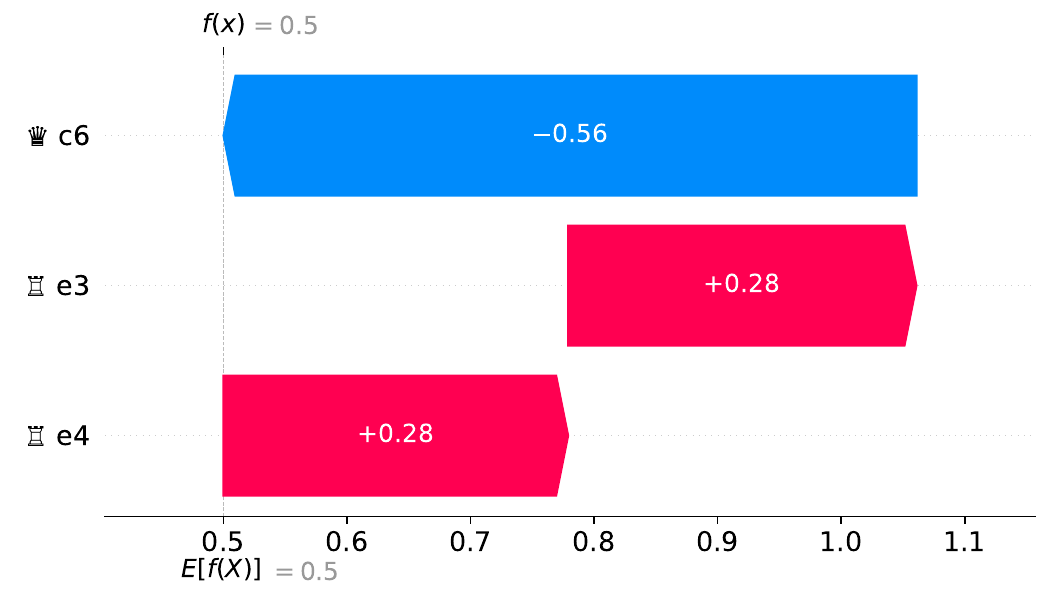}
  \end{subfigure}
  \caption{Explanation for the position in \Cref{fig:perturbations}. The two white rooks exactly offset the impact of the black queen, resulting in a perfectly drawn position.}
  \label{fig:example}
\end{figure}

A major challenge in explaining chess engine evaluations lies in the representation of the output. Most engines produce a scalar value in centipawns, where positive values indicate an advantage for White and negative values for Black. However, these scores are unbounded and require special handling for extreme cases such as forced mates, which are often represented by arbitrarily large constants. This lack of boundedness makes such outputs problematic for attribution methods like SHAP, which rely on finite and continuous model outputs to compute meaningful feature contributions.

To address this issue, we convert the centipawn evaluation into a probabilistic score, effectively framing the engine as a binary classifier that outputs the probability of a win for White. This transformation makes the output bounded in $[0,1]$ and aligns well with SHAP’s assumptions. Following the Lichess convention, we apply a logistic mapping from the centipawn score $s$ to a probability $p$:

\begin{equation}
    p(s) = \frac{1}{1 + \exp(-\beta s)},
\end{equation}

\noindent where $\beta \approx 3.68 \times 10^{-3}$ is a calibration parameter. This maps $s = 0$ to $p = 0.5$, representing equal chances for both players, and ensures smooth asymptotic behavior for large centipawn values.

In this setup, we reframe the chess engine as a black-box classifier $f: X \rightarrow [0,1]$ that maps a chess position $x \in X$ to the predicted probability that White will win the game. A position $x$ is represented as a list of individual chess pieces currently on the board, where each piece is characterized by its type (e.g., rook, knight), color (white or black), and square location. 
In addition to the piece list, a complete position includes auxiliary metadata such as the side to move (White or Black), castling rights, and en passant availability, all of which are required for engine evaluation.

Our goal is to explain the prediction $f(x)$ by quantifying the marginal contribution of each individual piece to the overall evaluation. To this end, we adapt SHAP (SHapley Additive exPlanations)~\cite{lundberg2017unified}, a model-agnostic interpretability framework that decomposes $f(x)$ into additive attributions assigned to each feature, to our purpose. In our context, we view chess pieces as features, and exploit SHAP to express the predicted outcome as a sum of contributions from each piece present on the board.

Formally, let $x'$ denote the set of non-king pieces present in the position $x$. Each element of $x'$ corresponds to a specific piece instance, uniquely defined by its type, color, and square. We restrict our SHAP analysis to features in $x'$ because removing either king would always result in an invalid position. Consequently, during SHAP perturbations, we hold both kings fixed and consider only the $n = |x'|$ non-king pieces as features. For any subset $S \subseteq x'$, we define $x_S$ as the perturbed position containing the pieces in $S$ together with both kings and the original metadata (e.g., side to move, castling rights).

The SHAP value $\phi_i$ assigned to a piece $i \in x'$ is defined as:

\begin{equation}
    \phi_i(f, x) = \sum_{S \subseteq x' \setminus \{i\}} \frac{|S|!(n - |S| - 1)!}{n!} \left[ f(x_{S \cup \{i\}}) - f(x_S) \right],
\end{equation}

\noindent where $f(x_S)$ denotes the engine's evaluation of the perturbed position, without piece $i$. 
An example of such perturbations can be viewed in \Cref{fig:perturbations}, starting from the original position (top-left), and removing pieces until there are none, (bottom-right).
In general, SHAP explanations are defined over binary vectors $z' \in \{0,1\}^n$ indicating feature presence or absence, and take the additive form $g(z') = \phi_0 + \sum_{i=1}^{n} \phi_i z'_i$. In our case, since all pieces in $x'$ are present in the original position $x$, each $z'_i = 1$, and the decomposition simplifies to the fully additive expression:
\begin{equation}
    f(x) = \phi_0 + \sum_{i=1}^{n} \phi_i.
\end{equation}

\noindent where $\phi_0$ is the base value of the model when only the two kings are present. Since engines universally treat king-only positions as trivially drawn, $\phi_0 = 0.5$, corresponding to a neutral evaluation. This eliminates the need to estimate the baseline from data and grounds the explanation in a well-defined and interpretable configuration. An illustrative example of such an explanation, based on the base position from \Cref{fig:perturbations}, is presented in \Cref{fig:example}. On the left, the position is visualized with each piece colored according to its SHAP contribution: red indicates a positive impact on White’s winning probability, while blue indicates a contribution favoring Black. On the right, the same contributions are displayed numerically, showing how each individual piece shifts the evaluation from the base value $\phi_0 = \mathbb{E}[f(X)] = 0.5$, corresponding to a balanced position with only kings, to the full evaluation of the current position, $f(x) = 0.5$.

SHAP requires evaluating $f(x_S)$ for all subsets $S \subseteq x'$ of non-king pieces. However, not all such perturbed positions are legal or evaluable by a chess engine. In particular, configurations that result in illegal checks or impossible side-to-move conditions may be rejected. To ensure that most inputs to SHAP are valid, we implement a carefully designed perturbation strategy. First, as explained above, we never ablate the kings, thereby guaranteeing that each perturbed position includes exactly one white king and one black king. In cases where a perturbed position is deemed illegal by the engine, such as when a side is giving checkmate and it is their turn to move, we attempt to restore legality by flipping the board, i.e., switching the side to move. This often resolves inconsistencies introduced by the ablation process. If the position remains invalid even after this adjustment, we assign it a default evaluation of $f(x_S) = 0.5$, corresponding to a draw. This fallback ensures that SHAP’s attribution remains well-defined while minimizing the introduction of bias or discontinuity in the output.

In summary, this approach yields an interpretable explanation of the engine's evaluation by attributing to each individual piece its marginal contribution to the predicted win probability for White. These attributions reveal how the presence of each piece increases or decreases the evaluation, providing intuitive insights into the strategic role each element plays in the position.

\section{Thematic Examples}
We show with thematic examples some positions in which SHAP can provide a good assessment of the pieces in the position, and some pitfalls showing limitations of such an approach. We use Stockfish 17.1 as the main engine, with a 5-second limit to evaluate the starting position, and a 0.1-second limit to evaluate the perturbations. We adopt the SHAP SamplingExplainer~\cite{lundberg2017unified}, evaluating a maximum of $10000$ perturbations per board.

\paragraph{Self-blocking Pawn.}
\definecolor{shapg1}{HTML}{053061}
\definecolor{shapf3}{HTML}{CA4942}
\definecolor{shapf4}{HTML}{A4CEE3}
\definecolor{shape6}{HTML}{EBF1F4}
\definecolor{shapg6}{HTML}{D2E5F0}
\definecolor{shapd7}{HTML}{3581B9}
\definecolor{shape7}{HTML}{D9E9F1}
\definecolor{shapg7}{HTML}{D5E7F0}
\definecolor{shapa8}{HTML}{870A24}

\begin{figure}[h]
  \centering
  \begin{subfigure}[c]{0.38\textwidth}
    \centering
  \resizebox{\linewidth}{!}{%
    \newgame
    \chessboard[
      setfen=Q7/3rpkp1/4p1p1/8/2K2P2/5R2/8/6q1 w - - 0 1,
      boardfontencoding=LSBC1,
      pgfstyle=color,
      opacity=0.7,
      color=shapg1,
  colorbackfield={g1},
  color=shapf3,
  colorbackfield={f3},
  color=shapf4,
  colorbackfield={f4},
  color=shape6,
  colorbackfield={e6},
  color=shapg6,
  colorbackfield={g6},
  color=shapd7,
  colorbackfield={d7},
  color=shape7,
  colorbackfield={e7},
  color=shapg7,
  colorbackfield={g7},
  color=shapa8,
  colorbackfield={a8}
    ]
  }
  \end{subfigure}%
  \hfill
  \begin{subfigure}[c]{0.62\textwidth}
    \centering
    \includegraphics[width=\linewidth]{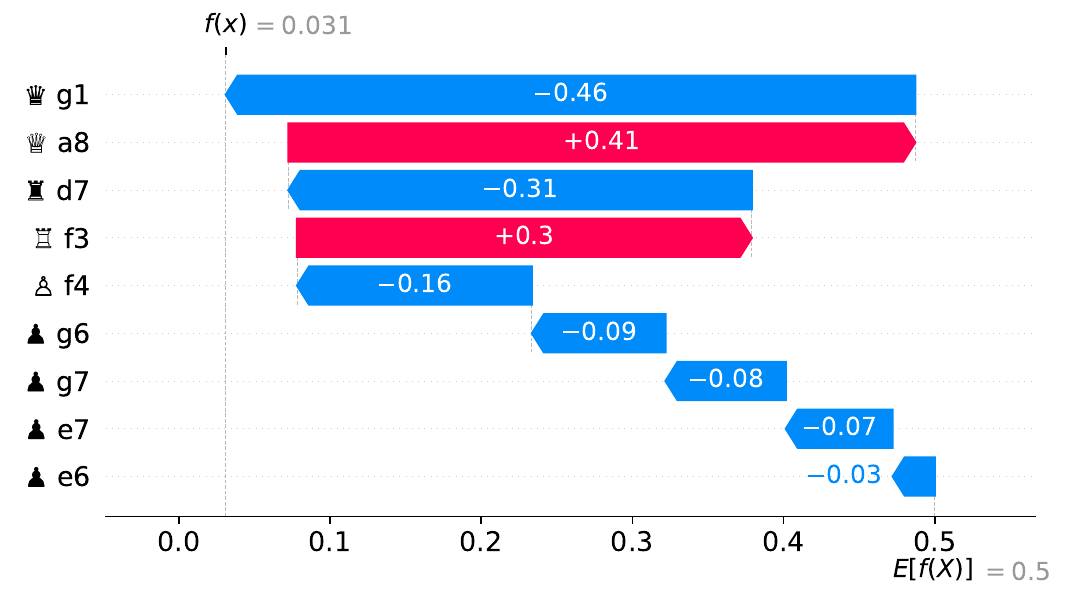}
  \end{subfigure}
  \caption{\textbf{Self-blocking pawn.} The position is completely winning for Black. Interestingly, the white pawn on \textbf{f4} aids Black’s position: without it, White would mate in several perturbations of the board.}
  \label{fig:self_blocking_pawn}
\end{figure}

One particularly compelling use case of the proposed explanations is their ability to highlight elements that are counterproductive to one's own position. In the position shown in \Cref{fig:self_blocking_pawn}, Black has a completely winning advantage. Remarkably, the white pawn on \textbf{f4}, rather than supporting White’s position, actually favors Black. In certain variations, the absence of this pawn would enable White to deliver checkmate, but its presence obstructs such tactical opportunities. This example illustrates how seemingly minor details can carry significant tactical weight. Even if such motifs are not immediately exploitable in the current game, recognizing them fosters deeper understanding and enhances pattern recognition.

\paragraph{Bishop vs Knight Endgame.}

\definecolor{shaph2}{HTML}{EF9B7A}
\definecolor{shapb3}{HTML}{EE9878}
\definecolor{shape3}{HTML}{67001F}
\definecolor{shapa4}{HTML}{ED9676}
\definecolor{shapg4}{HTML}{E6856A}
\definecolor{shapa5}{HTML}{ABD2E5}
\definecolor{shapg5}{HTML}{9DCAE1}
\definecolor{shapd6}{HTML}{2D76B4}
\definecolor{shape6}{HTML}{C2DDEB}
\definecolor{shaph6}{HTML}{B8D8E8}

\begin{figure}[h]
  \centering
  \begin{subfigure}[c]{0.38\textwidth}
    \centering
  \resizebox{\linewidth}{!}{%
    \newgame
    \chessboard[
      setfen=8/3k4/3np2p/p5p1/P5P1/1P2B3/2K4P/8 w - - 0 1,
      boardfontencoding=LSBC1,
      pgfstyle=color,
      opacity=0.7,
      color=shaph2,
  colorbackfield={h2},
  color=shapb3,
  colorbackfield={b3},
  color=shape3,
  colorbackfield={e3},
  color=shapa4,
  colorbackfield={a4},
  color=shapg4,
  colorbackfield={g4},
  color=shapa5,
  colorbackfield={a5},
  color=shapg5,
  colorbackfield={g5},
  color=shapd6,
  colorbackfield={d6},
  color=shape6,
  colorbackfield={e6},
  color=shaph6,
  colorbackfield={h6}
    ]
  }
  \end{subfigure}%
  \hfill
  \begin{subfigure}[c]{0.62\textwidth}
    \centering
    \includegraphics[width=\linewidth]{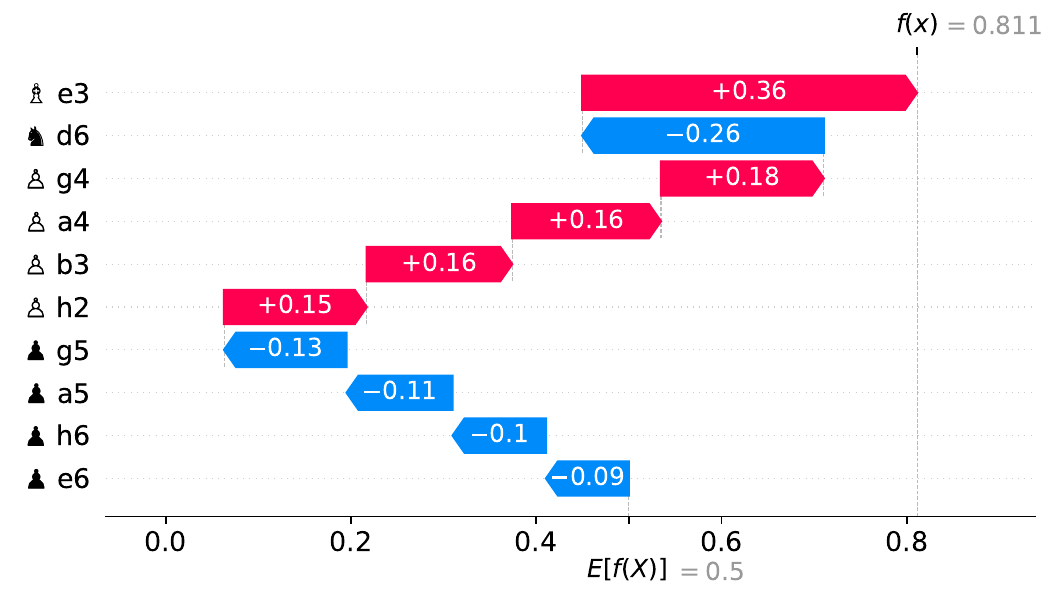}
  \end{subfigure}
  \caption{\textbf{Bishop vs Knight Endgame.} The bishop easily outmaneuvers the knight by switching wings, confirming its superiority in open endgames.}
  \label{fig:good_bishop}
\end{figure}

\revision{In the endgame puzzle in \Cref{fig:good_bishop}, White uses the bishop’s long range to outmaneuver the knight, shifting play from one side to the other until the knight falls behind: \variation{1. Bb6 Nb7 2. Bd4 Nd6 3. Bg7 Nf7 4. Bc3}.
White switches wings repeatedly, and the knight cannot keep pace; in the end, Black’s a-pawn is lost. SHAP correctly attributes greater importance to the bishop, which is generally stronger in endgames.}

\paragraph{Good Knight, Bad Bishop.} 
\definecolor{shape3}{HTML}{FACCB4}
\definecolor{shapg3}{HTML}{FACAB1}
\definecolor{shape4}{HTML}{4694C4}
\definecolor{shapg4}{HTML}{3D8BBF}
\definecolor{shapc5}{HTML}{D15749}
\definecolor{shape5}{HTML}{053061}
\definecolor{shapf5}{HTML}{559EC9}
\definecolor{shape7}{HTML}{C7433F}

\begin{figure}[h]
  \centering
  \begin{subfigure}[c]{0.38\textwidth}
    \centering
  \resizebox{\linewidth}{!}{%
    \newgame
    \chessboard[
      setfen=8/4B3/1K6/2Pknp2/4p1p1/4P1P1/8/8 b - - 0 133,
      boardfontencoding=LSBC1,
      pgfstyle=color,
      opacity=0.7,
      color=shape3,
  colorbackfield={e3},
  color=shapg3,
  colorbackfield={g3},
  color=shape4,
  colorbackfield={e4},
  color=shapg4,
  colorbackfield={g4},
  color=shapc5,
  colorbackfield={c5},
  color=shape5,
  colorbackfield={e5},
  color=shapf5,
  colorbackfield={f5},
  color=shape7,
  colorbackfield={e7}
    ]
  }
  \end{subfigure}%
  \hfill
  \begin{subfigure}[c]{0.62\textwidth}
    \centering
    \includegraphics[width=\linewidth]{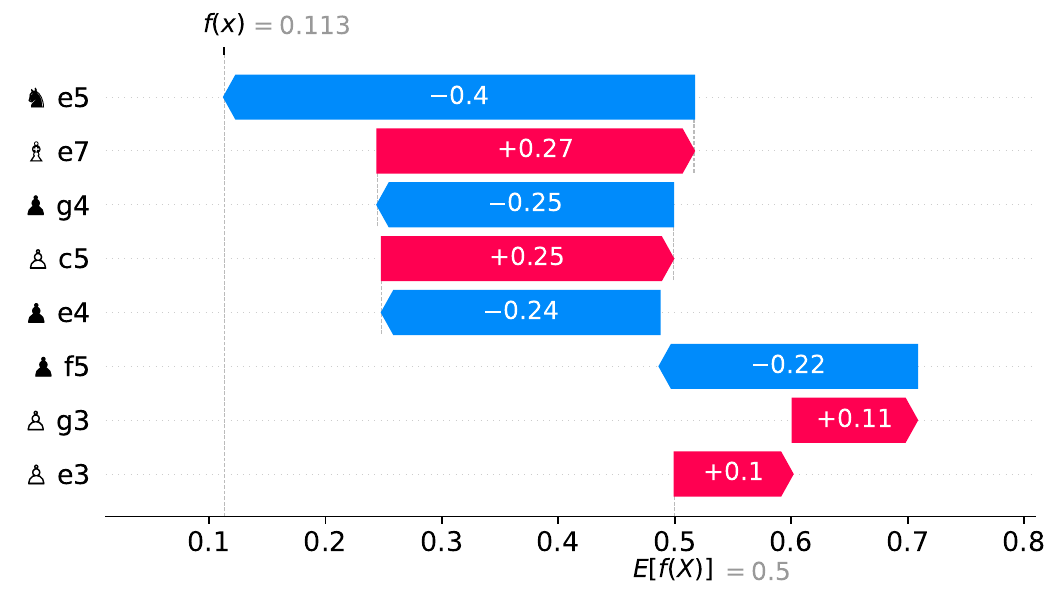}
  \end{subfigure}
  \caption{\textbf{Good Knight, Bad Bishop.} The knight dominates the board due to its flexibility and harmony with the pawn structure, while the bishop is severely restricted by its own pawns.}
  \label{fig:bad_bishop}
\end{figure}

\revision{Contrary to the previous example,} in the position shown in \Cref{fig:bad_bishop}, taken from the game Melkumyan–Gabuzyan, ARM-ch rapid, Yerevan 2016, the strategic inferiority of the bishop compared to the knight becomes evident. The critical pawn thrust \variation{133...f4!} initiates a dynamic transformation of the position, exposing the bishop's limited mobility and poor coordination with its own pawns. The resulting structure confines the bishop to passive squares, severely reducing its influence. In contrast, the knight demonstrates superior maneuverability, exploiting both color complexes and coordinating effectively with the advancing pawns. This imbalance ultimately leads to a decisive advantage for Black, highlighting the practical superiority of the knight in positions where the bishop is obstructed by its own pawn formation. This superiority is correctly highlighted by SHAP and confirmed through the concrete evaluation of the resulting endgame scenario.

\paragraph{Trapped Rook.}
\definecolor{shapb2}{HTML}{67001F}
\definecolor{shapa3}{HTML}{F8BFA3}
\definecolor{shapb3}{HTML}{F8F0EC}
\definecolor{shape4}{HTML}{F7B99B}
\definecolor{shapa5}{HTML}{D5E7F0}
\definecolor{shapb5}{HTML}{4C98C6}
\definecolor{shapd5}{HTML}{FBD0B9}
\definecolor{shape5}{HTML}{E1ECF3}
\definecolor{shapb6}{HTML}{EBF1F4}
\definecolor{shapd6}{HTML}{D9E9F1}
\begin{figure}[h]
  \centering
  \begin{subfigure}[c]{0.38\textwidth}
    \centering
  \resizebox{\linewidth}{!}{%
    \newgame
    \chessboard[
      setfen=8/4k3/1p1p4/pr1Pp3/4P3/PP1K4/1R6/8 w - - 0 31,
      boardfontencoding=LSBC1,
      pgfstyle=color,
      opacity=0.7,
      color=shapb2,
  colorbackfield={b2},
  color=shapa3,
  colorbackfield={a3},
  color=shapb3,
  colorbackfield={b3},
  color=shape4,
  colorbackfield={e4},
  color=shapa5,
  colorbackfield={a5},
  color=shapb5,
  colorbackfield={b5},
  color=shapd5,
  colorbackfield={d5},
  color=shape5,
  colorbackfield={e5},
  color=shapb6,
  colorbackfield={b6},
  color=shapd6,
  colorbackfield={d6}
    ]
  }
  \end{subfigure}%
  \hfill
  \begin{subfigure}[c]{0.62\textwidth}
    \centering
    \includegraphics[width=\linewidth]{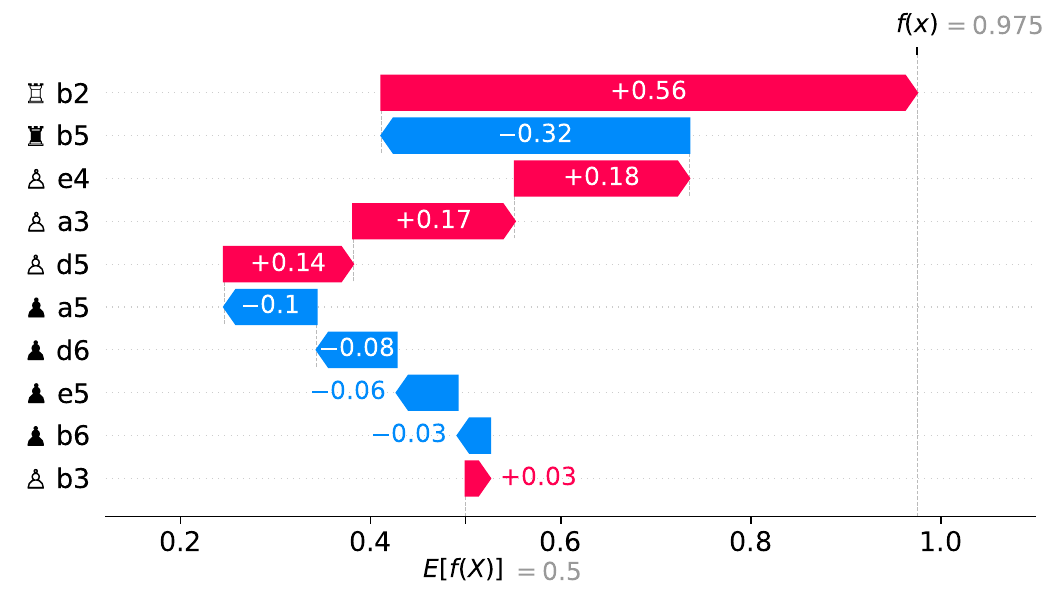}
  \end{subfigure}
  \caption{\textbf{Trapped rook.} The SHAP explanation highlights how the white rook is much more valuable than the black one.}
  \label{fig:trapped_rook}
\end{figure}

\Cref{fig:trapped_rook} illustrates an example of how positional constraints can drastically affect the evaluation of seemingly equivalent pieces. In this position, the black rook on \textbf{b5} contributes only $0.32$ towards Black's advantage, while the white rook on \textbf{b2} is valued at $0.56$. Despite both being rooks, the discrepancy arises from their differing mobility impact. The black rook is severely restricted in its movement due to surrounding pawns and limited open files, reducing its tactical and positional influence. In contrast, the white rook enjoys greater freedom and exerts pressure along key files.

\paragraph{Pins.}

    \definecolor{shapf3}{HTML}{C2383A}
\definecolor{shapa7}{HTML}{77B4D5}
\definecolor{shapb7}{HTML}{95C6DF}
\definecolor{shapc7}{HTML}{67001F}
\begin{figure}[htbp]
  \centering
  \begin{subfigure}[c]{0.38\textwidth}
    \centering
  \resizebox{\linewidth}{!}{%
    \newgame
    \chessboard[
      setfen=k7/rqQ3K1/8/8/8/5B2/8/8 w - - 0 1,
      boardfontencoding=LSBC1,
      pgfstyle=color,
      opacity=0.7,
      color=shapf3,
  colorbackfield={f3},
  color=shapa7,
  colorbackfield={a7},
  color=shapb7,
  colorbackfield={b7},
  color=shapc7,
  colorbackfield={c7}
    ]
  }
  \end{subfigure}%
  \hfill
  \begin{subfigure}[c]{0.62\textwidth}
    \centering
    \includegraphics[width=\linewidth]{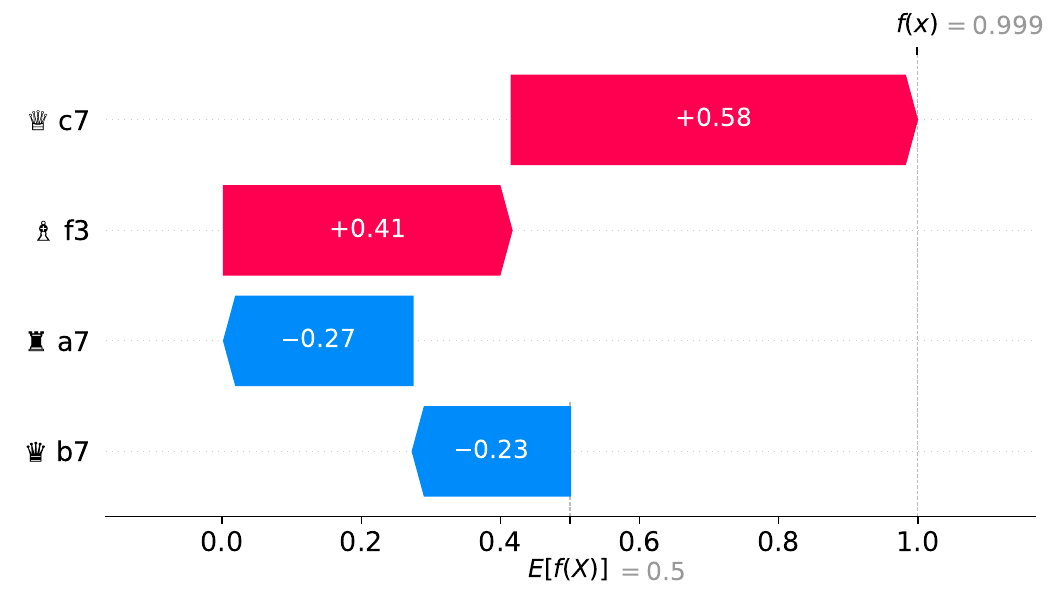}
  \end{subfigure}
  \caption{\textbf{Pins.} The high evaluation of the white bishop can help in identifying the two pins in this position.}
  \label{fig:tactic2}
\end{figure}

Explanations can also assist in identifying pinned pieces and the pieces responsible for the pin. In \Cref{fig:tactic2}, the black queen is evaluated as the worst piece, while the white bishop is ranked as the second best. This information can guide a novice player to recognize that the black queen is pinned by the white bishop, an extremely important piece in this configuration. Moreover, the similar evaluations of the white queen and bishop suggest that the white queen is not as valuable as it should be, as it is also pinned, indicating that either the bishop or the king is probably the piece to move. In this case, the best continuation is given by \variation{1. Kh8 Qxf3 2. Qc8#}.

\paragraph{Non-contributing Piece.}
Explanations can be used in chess puzzles to quickly identify pieces that contribute less to the position. For example, in \Cref{fig:tactic1}, the white knight plays a minimal role in White’s winning strategy, allowing the player to focus on the more critical pieces. In this case, the quickest mate is \variation{1. Qb4 Rxa7 2. Rc8#}.

    \definecolor{shape1}{HTML}{67001F}
\definecolor{shapa7}{HTML}{F9EDE7}
\definecolor{shapb7}{HTML}{FBD2BC}
\definecolor{shapc7}{HTML}{B82431}
\definecolor{shapa8}{HTML}{7DB8D7}
\begin{figure}[htbp]
  \centering
  \begin{subfigure}[c]{0.38\textwidth}
    \centering
  \resizebox{\linewidth}{!}{%
    \newgame
    \chessboard[
      setfen=rk2K3/NPR5/8/8/8/8/8/4Q3 w - - 0 1,
      boardfontencoding=LSBC1,
      pgfstyle=color,
      opacity=0.7,
      color=shape1,
  colorbackfield={e1},
  color=shapa7,
  colorbackfield={a7},
  color=shapb7,
  colorbackfield={b7},
  color=shapc7,
  colorbackfield={c7},
  color=shapa8,
  colorbackfield={a8}
    ]
  }
  \end{subfigure}%
  \hfill
  \begin{subfigure}[c]{0.62\textwidth}
    \centering
    \includegraphics[width=\linewidth]{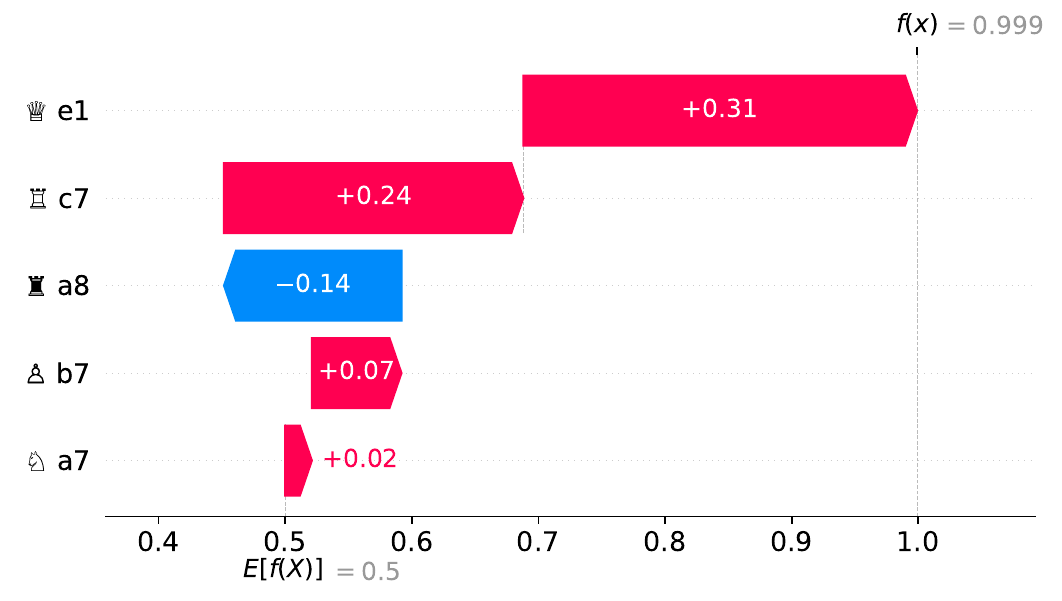}
  \end{subfigure}
  \caption{\textbf{Non-contributing Piece.} When analyzing a chess puzzle, it could be helpful for finding a solution to ignore pieces that do not contribute significantly to the position, in this case, the white knight.}
  \label{fig:tactic1}
\end{figure}

\paragraph{Comparing Engines.}

\definecolor{shape1}{HTML}{67001F}
    \definecolor{shapa2}{HTML}{FBE3D6}
\definecolor{shapf2}{HTML}{FBE0D0}
\definecolor{shapb3}{HTML}{F7B799}
\definecolor{shapg3}{HTML}{FAE5D9}
\definecolor{shapf4}{HTML}{67001F}
\definecolor{shapa5}{HTML}{D5E7F0}
\definecolor{shapc5}{HTML}{EEF3F5}
\definecolor{shapf6}{HTML}{C2383A}
\definecolor{shapg6}{HTML}{D5E7F0}
\definecolor{shaph6}{HTML}{FBD4BE}
\definecolor{shapd7}{HTML}{F0F3F5}
\definecolor{shape7}{HTML}{8AC0DB}
\definecolor{shapf7}{HTML}{D8E8F1}
\definecolor{shaph7}{HTML}{E5EEF3}
\definecolor{shapf8}{HTML}{9AC9E0}
\definecolor{shaph8}{HTML}{65A8CE}

\begin{figure}[htbp]
  \centering

  \begin{subfigure}[t]{\textwidth}
    \centering
    \begin{subfigure}[c]{0.38\textwidth}
      \centering
      \resizebox{\linewidth}{!}{%
        \newgame
        \chessboard[
          setfen=5rkq/3prp1p/5RpP/p1p5/5Q2/1B4P1/P4PK1/8 w - - 0 1,
          boardfontencoding=LSBC1,
          pgfstyle=color,
          opacity=0.7,
          color=shapa2, colorbackfield={a2},
          color=shapf2, colorbackfield={f2},
          color=shapb3, colorbackfield={b3},
          color=shapg3, colorbackfield={g3},
          color=shapf4, colorbackfield={f4},
          color=shapa5, colorbackfield={a5},
          color=shapc5, colorbackfield={c5},
          color=shapf6, colorbackfield={f6},
          color=shapg6, colorbackfield={g6},
          color=shaph6, colorbackfield={h6},
          color=shapd7, colorbackfield={d7},
          color=shape7, colorbackfield={e7},
          color=shapf7, colorbackfield={f7},
          color=shaph7, colorbackfield={h7},
          color=shapf8, colorbackfield={f8},
          color=shaph8, colorbackfield={h8}
        ]
      }
    \end{subfigure}%
    \hfill
    \begin{subfigure}[c]{0.62\textwidth}
      \centering
      \includegraphics[width=\linewidth]{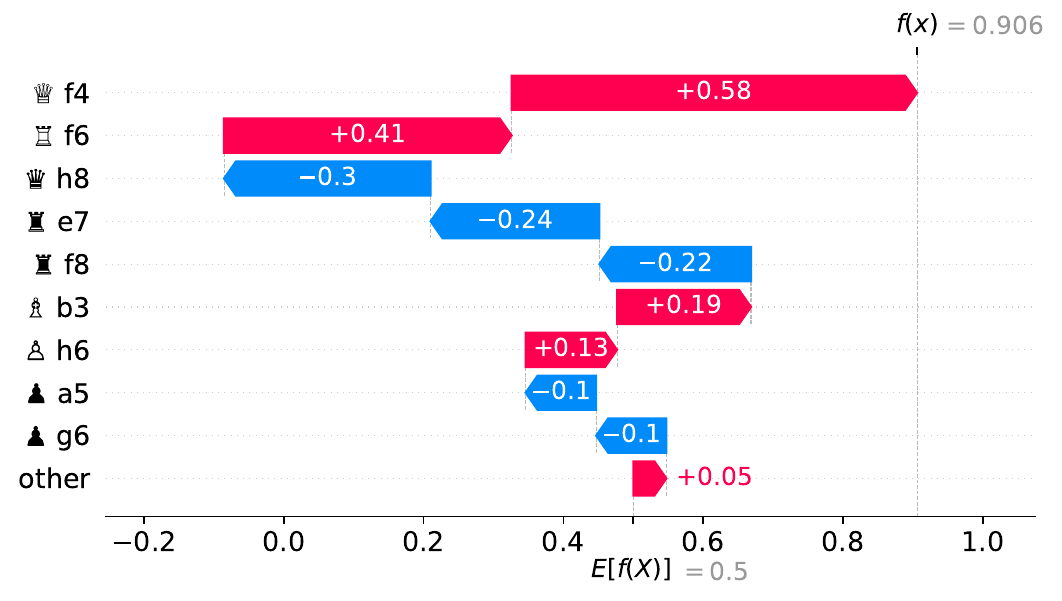}
    \end{subfigure}
    \caption{Stockfish}
  \end{subfigure}

  \vspace{1em}

  \begin{subfigure}[t]{\textwidth}
    \centering
    \begin{subfigure}[c]{0.38\textwidth}
      \centering
      \resizebox{\linewidth}{!}{%
        \newgame
        \chessboard[
          setfen=5rkq/3prp1p/5RpP/p1p5/5Q2/1B4P1/P4PK1/8 w - - 0 1,
          boardfontencoding=LSBC1,
          pgfstyle=color,
          opacity=0.7,
          color=shapa2, colorbackfield={a2},
          color=shapf2, colorbackfield={f2},
          color=shapb3, colorbackfield={b3},
          color=shapg3, colorbackfield={g3},
          color=shapf4, colorbackfield={f4},
          color=shapa5, colorbackfield={a5},
          color=shapc5, colorbackfield={c5},
          color=shapf6, colorbackfield={f6},
          color=shapg6, colorbackfield={g6},
          color=shaph6, colorbackfield={h6},
          color=shapd7, colorbackfield={d7},
          color=shape7, colorbackfield={e7},
          color=shapf7, colorbackfield={f7},
          color=shaph7, colorbackfield={h7},
          color=shapf8, colorbackfield={f8},
          color=shaph8, colorbackfield={h8}
        ]
      }
    \end{subfigure}%
    \hfill
    \begin{subfigure}[c]{0.62\textwidth}
      \centering
      \includegraphics[width=\linewidth]{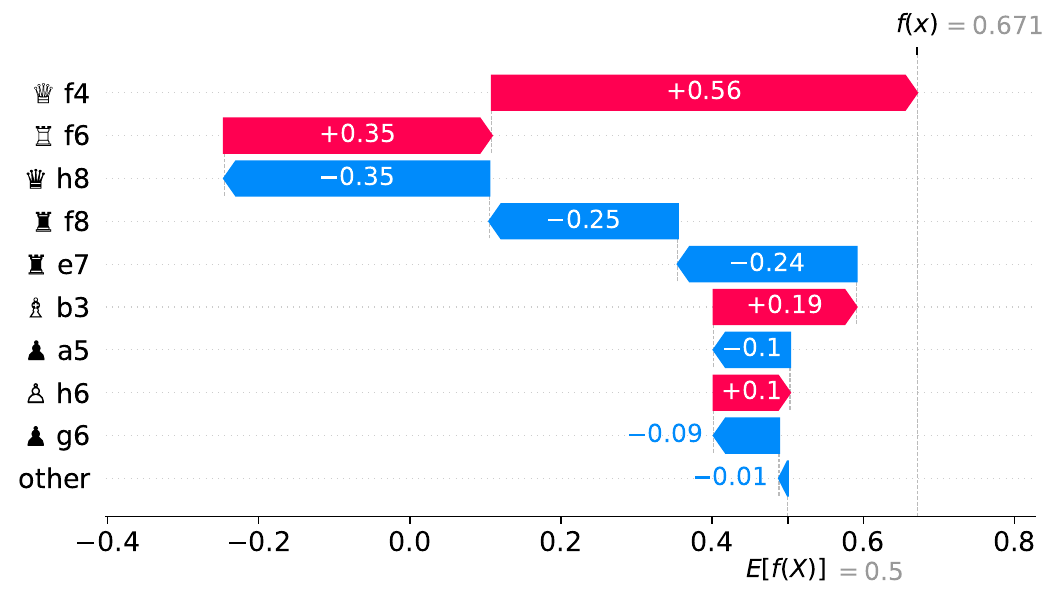}
    \end{subfigure}
    \caption{Leela}
  \end{subfigure}

  \caption{\textbf{Comparing Engines.} The same position can be evaluated differently by various engines. Explanations can help identify the pieces whose evaluations differ the most.}
  \label{fig:engines}
\end{figure}

Explanations can also be employed to compare different engine evaluations. In \Cref{fig:engines}, we examine the assessments made by Stockfish and Leela Zero (v0.31.2) on a critical position from the third game between Stockfish and AlphaZero (2018). While both engines agree that White stands better, they diverge significantly in their evaluation of the relative importance of individual pieces. The greatest disagreement concerns the white rook on \textbf{f6} and the black queen on \textbf{h8}: Stockfish assigns significantly higher value to the rook, whereas Leela regards the two pieces as similarly important. Such evaluations can be useful for understanding the strategic priorities and heuristics of different engines, and for uncovering subtleties that might otherwise be overlooked.

\subsection{Pitfalls}
Despite the interpretability benefits of SHAP-based attributions in the chess domain, several caveats must be considered to avoid misinterpreting the resulting scores. A central limitation is that SHAP values represent \textit{average marginal contributions }over all possible subsets (coalitions) of pieces. This means that a piece's attribution reflects its contribution in the context of many different board configurations, not just the one currently under analysis. Consequently, the attribution assigned to a piece does not directly correspond to the change in evaluation that would result from its removal. In other words, SHAP explanations capture statistical relevance across hypothetical perturbations rather than causal or deterministic influence in the original position.
Another key limitation is that the explanations are not guaranteed to be actionable. Many of the perturbed positions used in the SHAP computation may not be legally reachable from the original game state, due to the combinatorial nature of piece ablations and the disregard for move history. Therefore, while the model assigns attribution scores based on its evaluation function, these scores do not imply that a given piece can or should be moved or removed to obtain a particular outcome. Instead, the attributions are best interpreted as pedagogical tools: they offer insights into the strategic value assigned to each piece by the engine, helping players develop intuition and enhance their positional understanding.

\paragraph{King's Importance.}
One inherent limitation of the proposed approach is that the king's strength cannot be directly evaluated, as engines are unable to assess positions in which the king is absent. In the critical position taken from Denis Khismatullin vs. Pavel Eljanov, European Individual Championship Jerusalem ISR (2005), shown in \Cref{fig:active_king}, the best move is \variation{44.Kg1!}. SHAP is unable to attribute importance to it, as its removal invalidates the position from the engine’s perspective. Nevertheless, SHAP highlights other strategically significant features, such as the passed black pawns on \textbf{d3} which, if absent, leads to a white mate in several perturbations of the board.

\definecolor{shapd1}{HTML}{D6604D}
\definecolor{shapc2}{HTML}{165190}
\definecolor{shapf2}{HTML}{FBE1D2}
\definecolor{shapg2}{HTML}{F8EEE8}
\definecolor{shapd3}{HTML}{B0D4E6}
\definecolor{shape3}{HTML}{F9ECE5}
\definecolor{shaph3}{HTML}{F9E9DF}
\definecolor{shapb6}{HTML}{EAF1F4}
\definecolor{shapc6}{HTML}{F7B99B}
\definecolor{shapd6}{HTML}{87BEDA}
\definecolor{shapg6}{HTML}{D6E7F1}
\definecolor{shapf7}{HTML}{BDDAEA}
\definecolor{shaph7}{HTML}{DFECF2}
\definecolor{shapf8}{HTML}{67001F}
\begin{figure}[htbp]
  \centering
  \begin{subfigure}[c]{0.38\textwidth}
    \centering
  \resizebox{\linewidth}{!}{%
    \chessboard[
      setfen=5Q2/5p1p/1pPr2p1/6k1/8/3pP2P/2q2PP1/3R1K2 w - - 4 44,
      boardfontencoding=LSBC1,
      pgfstyle=color,
      opacity=0.7,
      color=shapd1,
      colorbackfield={d1},
      color=shapc2,
      colorbackfield={c2},
      color=shapf2,
      colorbackfield={f2},
      color=shapg2,
      colorbackfield={g2},
      color=shapd3,
      colorbackfield={d3},
      color=shape3,
      colorbackfield={e3},
      color=shaph3,
      colorbackfield={h3},
      color=shapb6,
      colorbackfield={b6},
      color=shapc6,
      colorbackfield={c6},
      color=shapd6,
      colorbackfield={d6},
      color=shapg6,
      colorbackfield={g6},
      color=shapf7,
      colorbackfield={f7},
      color=shaph7,
      colorbackfield={h7},
      color=shapf8,
      colorbackfield={f8}
    ]
  }
  \end{subfigure}%
  \hfill
  \begin{subfigure}[c]{0.62\textwidth}
    \centering
    \includegraphics[width=\linewidth]{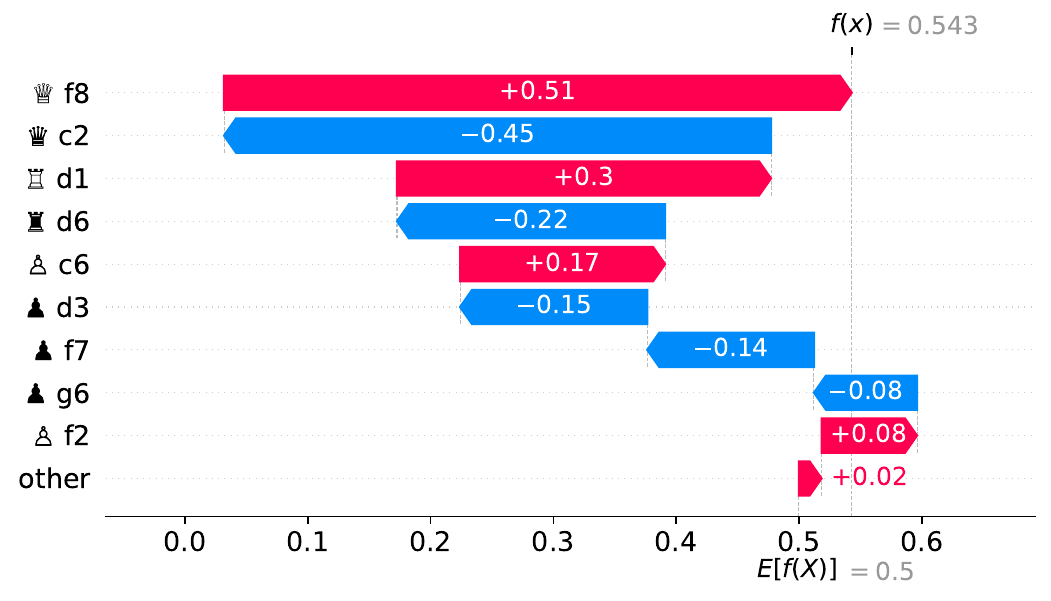}
  \end{subfigure}
  \caption{\textbf{Pitfall: King's Importance}. The best move in this position is \variation{44.Kg1}, however, an inherent limitation of the approach is that it is not able to assign importance to the king.}
  \label{fig:active_king}
\end{figure}

\paragraph{High Number of Pieces.}
When the board contains too many pieces, evaluating all possible combinations becomes computationally infeasible. Consequently, the estimated marginal contributions may overlook relevant configurations. Since the number of potential ablations grows exponentially with the number of pieces ($2^n$), brute-force methods quickly become impractical. 
\revision{In practice, a manageable upper limit is around $14$ pieces (excluding kings), which requires approximately $5$ minutes of computation\footnote{Apple M1 Pro, 32GB of RAM, Sonoma OS.}.}
For highly populated positions, targeted optimizations would be required to intelligently reduce the search space, ensuring that the resulting explanations remain accurate and reliable.

In summary, while SHAP provides a powerful framework for interpreting model predictions, its application in chess evaluation comes with inherent limitations. These explanations should be viewed as heuristic insights rather than prescriptive guides for decision-making.

\section{Discussion and Conclusion}
We have presented a method for attributing a chess engine's evaluation to individual pieces on the board by adapting SHAP, a principled model-agnostic interpretability framework, to the structured domain of chess. Our approach frames the engine as a probabilistic evaluator and computes piecewise contributions through systematic ablations, yielding additive and locally faithful explanations. The resulting attributions not only align with established pedagogical insights but also provide a rigorous foundation for analyzing the strategic and tactical value of each piece in a given position.

\revision{
A central motivation of this work lies in its didactic potential. Our explanations may provide a bridge between human teaching practices and modern chess engines. While this paper does not yet provide a controlled study with human learners, future evaluations in instructional settings could assess whether such explanations improve chess understanding and skill acquisition.
Beyond game analysis, this framework also opens promising avenues. One concrete direction suggested by our results is the evaluation of chess puzzles. Since puzzle quality is often judged by elegance, difficulty, and the contribution of specific pieces to the solution, piecewise attributions could provide quantitative support for puzzle generation and ranking.
However, despite its interpretability benefits, future work is needed to scale the approach to more complex positions. One promising direction is the incorporation of hierarchical or structured coalitions of pieces, which could reduce the exponential search space without sacrificing fidelity. Similarly, sampling strategies guided by strategic priors, rather than uniform random ablations, may offer further efficiency gains.}

\revision{Beyond chess, this methodology could generalize to other domains in which models evaluate structured states based on multiple interacting components, such as turn-based strategy settings, as well as non-game domains like multi-agent simulations or complex decision environments. These domains could benefit from similar forms of structured, per-component explanation. For example, in a turn-based strategy game, one could ablate individual units or resources to quantify their marginal impact on the probability of victory, highlighting which assets or tactical elements are most decisive. Likewise, in a multi-agent simulation, selectively removing or altering a single agent’s behavior could reveal how cooperation, competition, or coordination among agents contributes to emergent outcomes.} 
By bridging model-agnostic interpretability with combinatorial structure, our work contributes a reusable blueprint for localized attribution in settings where understanding why a model prefers a particular configuration is just as important as the evaluation itself.

\begin{acknowledgments}
  This work has been partially supported by the Italian Project Fondo Italiano per la Scienza FIS00001966 ``MIMOSA'', by the PRIN 2022 framework project ``PIANO'' (Personalized Interventions Against Online Toxicity) under CUP B53D23013290006, by the European Community Horizon~2020 programme under the funding schemes ERC-2018-ADG G.A. 834756 ``XAI'', by the European Commission under the NextGeneration EU programme – National Recovery and Resilience Plan (Piano Nazionale di Ripresa e Resilienza, PNRR) Project: ``SoBigData.it – Strengthening the Italian RI for Social Mining and Big Data Analytics'' – Prot. IR0000013 –  Av. n. 3264 del 28/12/2021, M4C2 - Investimento 1.3, Partenariato Esteso PE00000013 - ``FAIR'' - Future Artificial Intelligence Research'' - Spoke 1 ``Human-centered AI'', and ``FINDHR'' that has received funding from the European Union's Horizon Europe research and innovation program under G.A. 101070212.
\end{acknowledgments}

\bibliography{biblio}

\end{document}